\documentclass[preprint,3p, 12pt, authoryear]{elsarticle}

\usepackage{amssymb}
\usepackage{amsmath}

\usepackage{lineno}

\usepackage{times}  
\usepackage{helvet}  
\usepackage{courier} 
\usepackage{url}  
\usepackage{graphicx} 
\usepackage{natbib} 
\usepackage{caption} 
\usepackage{graphicx}
\usepackage{algorithm}
\usepackage{algorithmic}
\usepackage{changes}
\usepackage{tabularx}
\usepackage{xcolor}        
\usepackage{amsmath}
\usepackage{caption}
\usepackage{subcaption}
\usepackage{amssymb}
\usepackage{wrapfig} 
\usepackage{multirow}
\usepackage{changes}
\usepackage{balance}
\usepackage{array}
\usepackage{lmodern}
\usepackage[T1]{fontenc}
\usepackage{bm}
\usepackage{enumitem}

\newcolumntype{P}[1]{>{\centering\arraybackslash}p{#1}}

\newcommand{\head}[1]{\textnormal{\textbf{#1}}}
\newcommand{\flower}[1]{\textcolor{blue}{#1}}
\newcommand{\leaf}[1]{\textcolor{orange}{#1}}
\newcommand{\fruit}[1]{\textcolor{teal}{#1}}
\newcommand{\stem}[1]{\textcolor{purple}{#1}}

\newcommand{\cfix}[1]{#1}

\journal{}

\begin{document}

\begin{frontmatter}



\title{Automatic Fused Multimodal Deep Learning for Plant Identification} 


\author[l3]{Alfreds Lapkovskis \corref{cor2}}
\ead{alfreds.lapkovskis@dsv.su.se}
\author[l3]{Natalia Nefedova \corref{cor2}}
\ead{natalianefedova01@gmail.com}
\author[l3]{Ali Beikmohammadi \corref{cor1}}
\ead{beikmohammadi@dsv.su.se}
\affiliation[l3]{organization={Department of Computer and Systems Sciences},
            addressline={Stockholm University}, 
            postcode={SE-164 25}, 
            state={Stockholm},
            country={Sweden}}
\cortext[cor1]{Corresponding Author}
\cortext[cor2]{Authors contributed equally}

\begin{abstract}
Plant classification is vital for ecological conservation and agricultural productivity, enhancing our understanding of plant growth dynamics and aiding species preservation. 
The advent of deep learning (DL) techniques has revolutionized this field by enabling autonomous feature extraction, significantly reducing the dependence on manual expertise. 
However, conventional DL models often rely solely on single data sources, failing to capture the full biological diversity of plant species comprehensively. 
Recent research has turned to multimodal learning to overcome this limitation by integrating multiple data types, which enriches the representation of plant characteristics. 
This shift introduces the challenge of determining the optimal point for modality fusion.
In this paper, we introduce a pioneering multimodal DL-based approach for plant classification with automatic modality fusion. Utilizing the multimodal fusion architecture search, our method integrates images from multiple plant organs—flowers, leaves, fruits, and stems—into a cohesive model. 
\cfix{To address the lack of multimodal datasets, we contributed Multimodal-PlantCLEF, a restructured version of the PlantCLEF2015 dataset tailored for multimodal tasks.} Our method achieves 82.61\% accuracy on 979 classes of \cfix{Multimodal-PlantCLEF}, surpassing state-of-the-art methods and outperforming late fusion by 10.33\%. Through the incorporation of multimodal dropout, our approach demonstrates strong robustness to missing modalities.
We validate our model against established benchmarks using standard performance metrics and McNemar's test, further underscoring its superiority.
\end{abstract}

\begin{keyword}


Plant Identification \sep Multimodal Learning  \sep Fusion Automation \sep Multimodal Fusion Architecture Search \sep Neural Architecture Search \sep Multimodal Dataset
\end{keyword}

\end{frontmatter}





\section{Introduction}
Plant classification is among the most significant tasks for agriculture and ecology, it facilitates the preservation of plant species and enhances understanding of their growth dynamics, thus protecting the environment \citep{Zhang_2012}. 
Typically, plants can be categorized into additional specific groupings, such as weeds, invasive species, and plants exhibiting diseases and conditions. 
According to \citet{Dyrmann_2016}, between 23 and 71 percent of yield can be lost due to uncontrolled weeds, highlighting the necessity of accurately understanding weed species for more precise herbicide application. 
Therefore, accurate plant
identification is crucial in preventing crop losses and avoiding inaccurate and unnecessary pesticide usage \citep{Meshram_2021}.

Manual plant identification typically relies on leaf and flower features and demands profound domain expertise, alongside substantial allocation of time and financial resources \citep{Beikmohammadi_2022}. 
Moreover, \citet{Saleem_2018} suggest that the extensive diversity of plant species amplifies the complexity of laboratory classification. Given these challenges, there has been a shift towards automated methods. 
These approaches, leveraging machine learning (ML) and computer vision \citep{Waldchen_2018}, aim to predict plant types and minimize the reliance on manual skill and resources. 

In this regard, some studies have explored traditional ML algorithms for plant classification \citep{Gao_2018,Saleem_2018}. However, these algorithms rely on the creation of hand-crafted features, a process heavily dependent on human expertise \citep{Haichen_2021}. 
This human involvement also introduces the potential for biased assumptions and poses challenges in manually identifying suitable features for visual classification.  
For instance, a classifier based on leaf teeth proves ineffective for species lacking prominent leaf teeth \citep{Waldchen_2018}. 
Similarly, classifiers relying on leaf contours struggle with species exhibiting similar leaf shapes \citep{Liu_2016}. 

Recognizing these challenges, numerous studies indicate the superior performance of deep learning (DL) 
models compared to traditional ones \citep{Nhan_2020,Kolhar_2021,Kaya_2019}.
Consequently, researchers have recently adopted DL techniques to develop more effective models for plant identification \citep{Beikmohammadi_2018, Beikmohammadi_2022, Espejo-Garcia_2020, Ghazi_2017, Ghosh_2022, Kaya_2019, Lee_2018, Liu_2022, Tan_2020}. 
These models, typically employing convolutional neural networks (CNNs), can extract features themselves without the need for explicit feature engineering \citep{Nhan_2020,Saleem_2018,Waldchen_2018}. However, DL models introduce challenges in engineering model architectures, a task that demands expertise and extensive experimentation and is susceptible to errors \citep{Elsken_2019}.
Addressing this, neural architecture search (NAS), which automates the design of neural architectures, has demonstrated remarkable performance, surpassing manually designed architectures across various ML tasks, notably in image classification \citep{Liu_2021}. These methods have also been effectively applied in plant identification \citep{Umamageswari_2023,Sun_2022}.

However, typically, both traditional ML and DL models developed for plant classification tasks are
constrained to a single data source, often leaf or whole plant images. From a biological standpoint, a single organ is insufficient for classification \citep{Nhan_2020}, as variations in appearance can occur within the same species due to various factors, while different species may exhibit similar features. 
Moreover, using a whole plant image is insufficient, as different organs vary in scale, and capturing all their details in a single image is impractical \citep{Waldchen_2018}. 
In response to this limitation, very recent studies have delved into the application of multimodal learning techniques \citep{deLutio_2021, Liu_2016,Nhan_2020, Salve_2018, Trong_2020, Wang_2022, Zhou_2021}, which integrate diverse data sources to provide a comprehensive representation of phenomena. 
Particularly, \citet{Nhan_2020} illustrate that leveraging images from multiple plant organs outperforms reliance on a single organ, in line with botanical insights \cfix{\citep{Goeau_2015}}. 
\citet{Waldchen_2018} underscore the emerging trend of multi-organ-based plant identification, indicating promising accuracy improvements due to the diverse plant viewpoints.

In multimodal learning, the fusion of modalities is recognized as a critical challenge
\citep{Barua_2023,Zhang_2020}.
Various fusion strategies outlined in the literature include early, intermediate, late, and hybrid fusions \citep{Boulahia_2021}. Early fusion integrates modalities before feature extraction, such as combining multiple 2D images into a single tensor. Intermediate fusion extracts features from each modality separately and then merges them, offering deeper insights. Late fusion combines modalities at the decision level, often through averaging. The hybrid approach mixes these strategies for optimal results.

\cfix{
Among these fusion strategies, late fusion emerges as the most prevalent in the observed plant classification literature, presumably due to its simplicity and adaptability \citep{Baltrusaitis_2019}. 
However, the choice of a specific strategy relies on the discretion of the model developer \citep{Xu_2021}, which can introduce bias and lead to a suboptimal architecture. 
}

\cfix{
To address these issues, we propose an automatic multimodal fusion approach utilizing images from four distinct plant organs---flowers, leaves, fruits, and stems. Following \citet{Nhan_2020}, we refer to these organs as \emph{modalities}. While all organs are represented as RGB images, each encapsulates a unique set of biological features, reflecting the fundamental property of multimodality--- \emph{complementarity} \citep{Lahat_2015}. 
Furthermore, unlike multi-view methods \citep{Seeland_2021}, our model explicitly requires distinct plant organs as inputs. Additionally, unlike certain multi-organ methods \citep{Lee_2018}, our model has a fixed set of inputs, with each input corresponding exclusively to a specific organ.
Thus, we suggest the term \emph{multimodality} aligns more accurately with our approach.
}

\cfix{
That said, existing plant classification datasets are predominantly designed for unimodal tasks, which poses a significant challenge for developing and evaluating multimodal approaches. To address this limitation, we introduce a data preprocessing pipeline that transforms an existing unimodal dataset, namely PlantCLEF2015 \citep{Joly_2015}, into a multimodal dataset comprising combinations of plant organ images. This dataset, which we call \emph{Multimodal-PlantCLEF}, supports the development of models with a fixed number of inputs, each corresponding to a specific plant organ.
}

\cfix{
Our contribution is fourfold:
}
\begin{enumerate}
    \item \cfix{We propose a novel data preprocessing approach to convert a unimodal plant classification dataset into a multimodal one and apply it to transform PlantCLEF2015 into Multimodal-PlantCLEF for use in this study.}

    \item \cfix{We propose a novel automatic fused multimodal DL approach for the first time in the context of plant classification. To do so, we first train a unimodal model for each modality using the MobileNetV3Small pre-trained model. Then, we apply a modified multimodal fusion architecture search algorithm (MFAS) \citep{Perez-Rua_2019} to automatically fuse these unimodal models.}

    \item \cfix{We evaluate the proposed model against an established baseline in the form of late fusion using an averaging strategy \citep{Baltrusaitis_2019}, utilizing standard performance metrics and McNemar's statistical test \citep{Dietterich_1998}. The results demonstrate that our automated fusion approach enables the construction of a more effective multimodal DL model for plant identification, outperforming the baseline.}

    \item \cfix{Additionally, we assess the proposed model on all subsets of plant organs, revealing its robustness to missing modalities when trained with a multimodal dropout technique \citep{Cheerla_2019}.}
\end{enumerate}

The source code of our work is available online at the GitHub repository\footnote{GitHub repository: https://github.com/AlfredsLapkovskis/MultimodalPlantClassifier}.

The remainder of this paper is organized as follows. Section \ref{section:methods} introduces our proposed method, outlines the dataset and data preprocessing, and describes model evaluation. Section \ref{section:results} presents the obtained results. Section \ref{section:discussion} discusses the acquired results in relation to other research. Section \ref{section:conclusion} concludes the paper and highlights the directions for future research.

\section{Materials and Methods} \label{section:methods}
\subsection{Fusion Automation Algorithm Selection}
Different layers of DL models represent different levels of abstractions, and the highest levels are not necessarily the most suitable for fusion \citep{Perez-Rua_2019}. 
Following this insight, to automate the development of a multimodal model architecture, we explore the use of NAS algorithms specifically designed for multimodal frameworks. 
These algorithms offer a promising solution for automatically identifying the optimal point of fusion rather than relying on manual determination.
In this regard, \citet{Perez-Rua_2019} introduce MFAS, a multimodal NAS algorithm. 
Central to their approach is the assumption that each modality possesses a distinct pre-trained model, substantially reducing the search space by maintaining these models static during the search process. 
The MFAS algorithm iteratively seeks an optimal joint architecture by progressively merging individual pre-trained models at different layers.
A notable advantage of this methodology lies in its focus on training fusion layers exclusively, resulting in significant computational time savings. 

In another work, \citet{Xu_2021} present MUFASA, an advanced multimodal NAS algorithm. One of the standout features of this algorithm is its comprehensive approach, which involves searching for optimal architectures not only for the entire fusion architecture but also for each modality individually, all while considering various fusion strategies. Unlike unimodal NAS methods, MUFASA addresses the whole architecture, leveraging the understanding of its multimodal nature. 
Furthermore, unlike \citeauthor{Perez-Rua_2019}'s \citeyearpar{Perez-Rua_2019} algorithm, MUFASA addresses the architectures of individual modalities while considering their interdependencies. 
This unique approach positions MUFASA as potentially more powerful at tackling challenges in multimodal fusion.

Among these algorithms, in this study, we select the approach proposed by \citeauthor{Perez-Rua_2019} \citeyearpar{Perez-Rua_2019}, as it is considered more suitable for efficiently searching for optimal multimodal fusion architectures. 
While MUFASA demonstrates superior potential, it comes with a notable drawback: its high computational demands. 
\citet{Xu_2021} indicate that achieving state-of-the-art performance on widely used academic datasets would necessitate roughly two CPU years of computational time. 
In contrast, \citeauthor{Perez-Rua_2019}'s \citeyearpar{Perez-Rua_2019} multimodal architecture search has achieved high accuracy, completing within 150 hours of four P100 GPU time on a large-scale image dataset and much faster on simpler datasets. 
This difference can be attributed, at least, to the substantial training requirements of MUFASA, which involves optimizing a larger number of weights and evaluating a greater number of configurations.

\subsection{Proposed Method}

The MFAS algorithm employed in this work requires a pre-trained unimodal model for each modality. Therefore, the method is initiated with the creation of these models.

\subsubsection{Unimodal Models} \label{section:unimodal_models}
To construct a unimodal model for each modality represented by different plant organs,
we employ a transfer learning technique. 
There are two primary approaches to applying transfer learning. 
One involves utilizing pre-trained model weights to extract features from a dataset for subsequent classification, while the other entails full or partial updates of pre-trained weights using a new dataset, known as fine-tuning \citep{Espejo-Garcia_2020}. Initially, we adopted the former approach, pre-training only the appended top layers of each unimodal model. For models where fine-tuning yielded improved performance, we selectively proceeded with this technique.

We employ MobileNetV3Small \citep{Howard_2019} as the base model, utilizing weights pre-trained on the ImageNet dataset, for our transfer learning approach.
This model is selected for its balance of depth and high performance, compatibility with RGB inputs, and design for an input size close to $256\times256\times3$.

\subsubsection{Multimodal Fusion Architecture Search}
We leverage the MFAS algorithm, which utilizes a pre-trained model $\bm{f}^{(i)}(\bm{x}^{(i)})=\hat{y_i}$ for each modality $i \in \{1, \dots, m\}$. Here, $\hat{y_i}$ denotes an approximation of the true label $y$, derived from the input $\bm{x}$ specific to modality $i$. 
Each model $\bm{f}^{(i)}$ with size $n_i$ is composed of layers $\bm{f}^{(i)}_l$, where $l \in \{1,\dots,n_i\}$, such that for $l$-th layer, $\bm{x}^{(i)}_l=(\bm{f}^{(i)}_l\circ \bm{f}^{(i)}_{l-1}\circ\dots \circ \bm{f}^{(i)}_1)(\bm{x}^{(i)})$ represents features considered for fusion with features from other modalities. 
The objective of the algorithm is to find optimal combinations of such features to fuse, along with determining the properties of such fusion.

To do so, fusion is performed through another model $\bm{h}$, whose layers are defined as follows:

\begin{equation}
    \bm{h}_1 = \bm{\sigma}_{\gamma^a_1}
    \begin{pmatrix}
        \bm{W}_1
        \begin{bmatrix}
            \bm{x}^{(1)}_{\gamma^1_1}\\
            \vdots\\
            \bm{x}^{(m)}_{\gamma^m_1}\\
        \end{bmatrix}
    \end{pmatrix},
    \bm{h}_l = \bm{\sigma}_{\gamma^a_l}
    \begin{pmatrix}
        \bm{W}_l
        \begin{bmatrix}
            \bm{x}^{(1)}_{\gamma^1_l}\\
            \vdots\\
            \bm{x}^{(m)}_{\gamma^m_l}\\
            \bm{h}_{l-1}\\
        \end{bmatrix}
    \end{pmatrix}, 
    l > 1
\end{equation}
\noindent where for $l$-th layer of $\bm{h}$, $\bm{\sigma}_{\gamma^a_l}$ denotes an activation function, $\bm{W}_l$ is a trainable weight matrix and $\gamma_l=(\gamma^1_l, \dots, \gamma^m_l, \gamma^a_l)$ is a tuple with indices of features from each modality and an index of an activation function. 
The maximum number of fusion layers $\bm{h}_l$ is denoted by $L: l \in \{1, \dots, L\}$, and is a hyperparameter defined prior to execution of the algorithm. 
A complete fusion configuration of a particular instance of $\bm{h}$ with $L$ layers is defined by a vector of tuples $[\gamma_l]_{l \in \{1, \dots, L\}}$, while a set of all possible tuples with $L$ layers is denoted by $\bm{\Gamma}_L$. 
A list of possible activation functions with size $k$ and possible modality layers used in $\bm{h}_l$ are also hyperparameters that can be implementation-specific. 

Given this setup, the algorithm spans a large search space of size $(n_1 \times \dots \times n_m \times k)^L$. 
Since exploring even a small portion of these configurations manually is infeasible, \citet{Perez-Rua_2019} have integrated a sequential model-based optimization (SMBO) method into their framework. 
This approach methodically explores the search space by progressively introducing new configurations, a process which has proven to yield architectures that perform comparably to those identified through direct methods \citep{Perez-Rua_2019}.

Although our approach to fusion automation is grounded in the methodologies outlined in \citep{Perez-Rua_2019}, it also incorporates several modifications while maintaining certain similarities. 
In the following, we provide further elaboration on the specific details of our implementation.

\begin{itemize}[wide, labelindent=0pt]
    
    \item[] \noindent \textbf{Unimodal models:} We utilize the pre-trained MobileNetV3Small model for each modality as detailed in Section \ref{section:unimodal_models}.
    
    \item[] \noindent \textbf{Search space:} In this paper, the search space is constrained to the following parameters:
    \begin{itemize}
        \item Maximum of four fusion layers (i.e., $L=4$).
        \item Two possible activation functions: ReLU and sigmoid (i.e., $k=2$).
        \item 6 fusible layers in each unimodal model: the 1st activation, the 1st, 6th, and 11th inverted residual blocks, and our two appended dense layers (i.e., $n_i=6, \forall i$). 
    \end{itemize}
    
    This results in $6^4\times2=2592$ different configurations for a single fusion layer, and forms a search space of size $2592^4\approx4.51\times10^{13}$. 

    \item[] \noindent \textbf{Generating model configurations:}
    In this work, the sampled architectures, denoted by \(\mathcal{S}\), are reused across iterations, in contrast to the approach taken by \citet{Perez-Rua_2019}, where architectures are regenerated in each iteration. During the progression from $l=1$ to $l=L$, newly generated layer configurations are either appended to the sampled configurations or used to replace existing $l$-th layers in $\mathcal{S}$.
    
    \item[] \noindent \textbf{Building architectures:} Multimodal architectures are constructed based on fusion configurations. Since the outputs of layers from unimodal models may have different numbers of dimensions, similar to \citet{Perez-Rua_2019}, global average pooling is applied to each multidimensional output from MobileNetV3Small.
    Subsequently, the output of each fused layer is concatenated and connected with a dense layer, incorporating an activation function specified in the model configuration. In the case of an intermediate fusion layer ($l>1$), the previous fusion layer is also connected to the dense layer. Each multimodal model is finalized with a classifier layer employing a softmax activation function.
    
    \item[] \noindent \textbf{Weight sharing:} In contrast to \citet{Perez-Rua_2019}, where weights are shared between fusion layers that have the same indices and identical weight matrix sizes, our approach also takes into account the activation functions of the fusion layers. This adjustment acknowledges the significant impact that activation functions can have on the behavior of weights. 
    In our model, weights are shared across all configurations.
    
    \item[] \noindent \textbf{Storing results:} Similar to
    \citet{Perez-Rua_2019}, if the same configuration is visited twice, the best result is retained.
    
    \item[] \noindent \textbf{Surrogate:} In our work, a surrogate similar to the one employed by \citet{Perez-Rua_2018} is utilized, as it has proven to be effective in this context. This surrogate comprises an embedding layer with zero masking, yielding vectors with a length of 100; an LSTM layer with 100 neurons; and a regression layer with a single neuron and sigmoid activation. The model is compiled with an Adam optimizer with a learning rate (LR) of 0.001, and mean squared error (MSE) loss. Each update is executed for 50 epochs with a batch size of 64.

    Despite \citet{Perez-Rua_2019}, where batches for the surrogate were created by grouping stored configurations based on their lengths, in this study, we address this issue by applying right zero padding to configurations.
    Moreover, while \citet{Perez-Rua_2019} suggest updating the surrogate model only with recently generated data, our approach diverges by updating the surrogate each time with the entire dataset of results.

    \item[] \noindent \textbf{Temperature-based sampling:} According to \citet{Perez-Rua_2018}, the probability of sampling an architecture $i$ with a score $a_i$ is $p_i=a_i/\sum_ja_j$. However, when incorporating a temperature factor $t$, the probability of sampling $i$ is given by $p_i^{1/t}/\sum_jp_j^{1/t}$. Thus, the larger the $t$, the more stochastic the sampling becomes. Similar to \citet{Perez-Rua_2019}, this research employs inverse exponential temperature scheduling, as it has shown to perform well in this context too. Therefore, at each step $s$ of the MFAS algorithm, the temperature is computed as $(t_{max}-t_{min})e^{-(s/d)^2}+t_{min}$, where $d$ is the decay rate.
\end{itemize}

\subsection{Dataset} \label{section:dataset}

\begin{figure} [H]
    \begin{subfigure}{0.5\textwidth}
        \centering
        \includegraphics[scale=0.5]{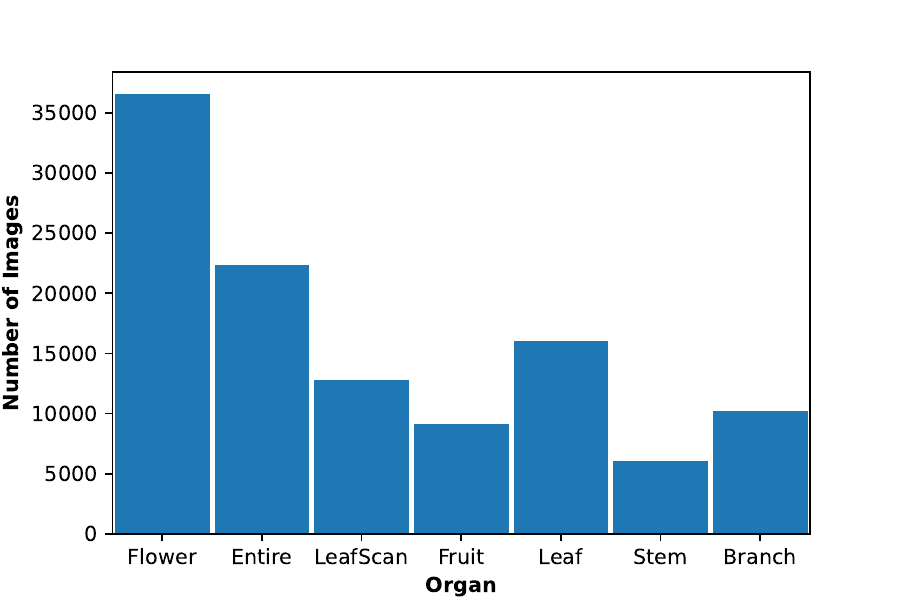}
        \caption{}
        \label{fig:dataset_organs}
    \end{subfigure}
    \begin{subfigure}{0.5\textwidth}
        \centering
        \includegraphics[scale=0.5]{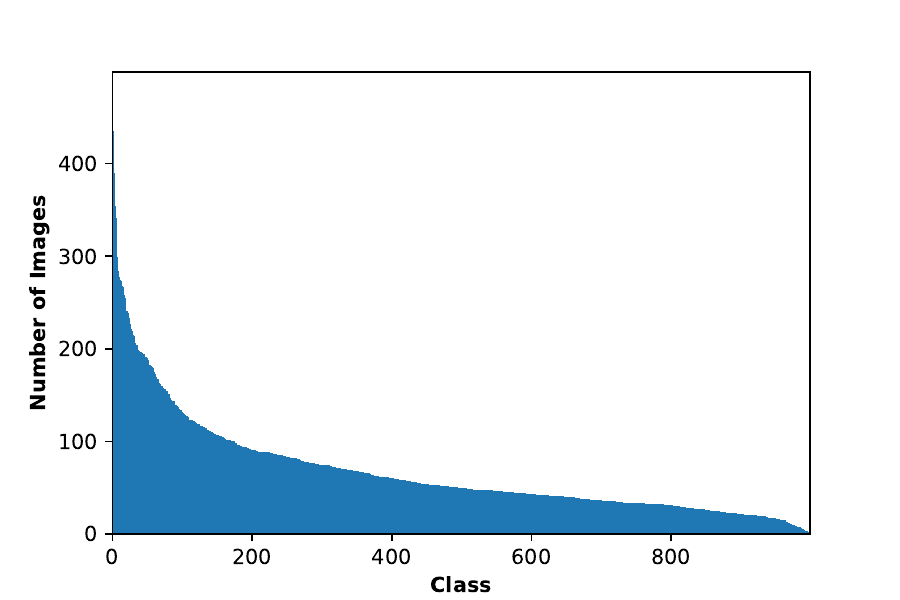}
        \caption{}
        \label{fig:class_dist}
    \end{subfigure}
    \caption{PlantCLEF2015 Dataset Summary}
    \small{(a) Distribution of plant organs in the dataset, (b) Class distribution in the dataset, considering only the relevant organs: flowers, leaves, fruits, and stems.}
\end{figure}

We selected the PlantCLEF2015 dataset \citep{Joly_2015} for our proposed plant classification model because it contains images of different plant organs---including flowers, leaves, fruits, and stems---providing a large and diverse collection that ensures a sufficient quantity of images for each organ type as depicted in Figure \ref{fig:dataset_organs}.

\begin{figure}[H]
    \centering
    \includegraphics[width=1\textwidth]{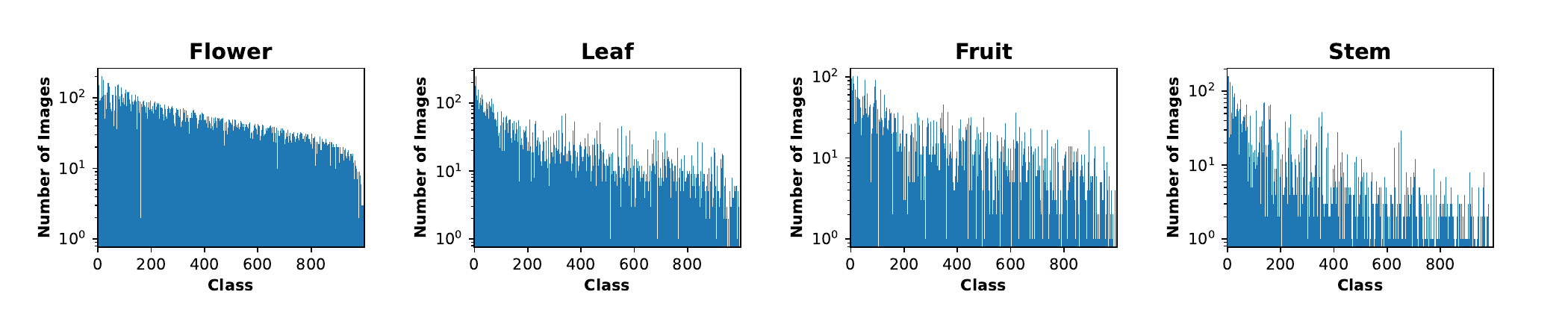}
    \caption{Distribution of Plant Organs among Classes in PlantCLEF2015}
    \label{fig:class_organ_dist}
\end{figure}

This dataset follows a long-tailed distribution, as depicted in Figure \ref{fig:class_dist}. This class imbalance in the data underscores the necessity of employing additional techniques to mitigate its potentially adverse effects on model learning. Furthermore, Figure \ref{fig:class_organ_dist} illustrates the distribution of organ images across classes. Notably, a considerable number of classes entirely lack certain organs. This observation highlights the necessity for models to handle missing modalities effectively, necessitating tailored training approaches.

Moreover, it is evident that flowers, being a discriminative organ \citep{Nhan_2020}, are available for most classes, followed by leaves and fruits, although with a noticeable scarcity in many classes. This scarcity is even more pronounced in stems, being the least discriminative and the least available organ across classes. 

\begin{figure}[H]
    \centering
    \includegraphics[width=1\textwidth]{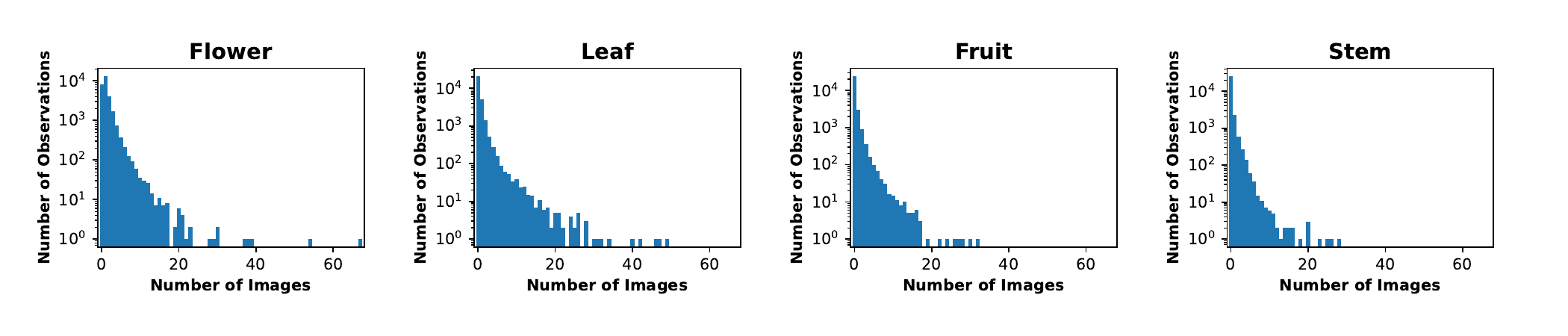}
    \caption{Distribution of Plant Organs among Observations in PlantCLEF2015}
    \label{fig:organ_dist}
\end{figure}

Finally, the images in the dataset are derived from observations---individual plant specimens. It is essential to use this information to split the data for training, validation, and evaluation based on observations rather than images, in order to prevent potential bias from exposing similar images of the same observation across different data splits. However, achieving well-balanced splits in this context is non-trivial, as each observation can have an arbitrary number of images, as illustrated in Figure \ref{fig:organ_dist}. Our approach to data splitting under these circumstances is detailed in Section \ref{section:preprocessing}.

\subsubsection{Data Preprocessing} \label{section:preprocessing}

\cfix{
The original PlantCLEF2015 dataset and its associated evaluation protocols are not suitable for the development of our proposed multimodal model. These protocols are designed for models that accept either a single image or an observation comprising an arbitrary number of images from various plant organs. In contrast, our model requires a fixed input format consisting of four specific plant organ images. This fundamental mismatch necessitates a custom strategy to reorganize the original dataset into fixed combinations of plant organ images, which inherently precludes the use of standard PlantCLEF2015 evaluation protocols. Given these constraints, modifying the dataset is both a justified and essential step.
It is important to note that all multimodal approaches are dealing with this issue in plant identification. For more information, see Section \ref{section:limitations}.
In the following, we detail our preprocessing pipeline for transforming PlantCLEF2015 into Multimodal-PlantCLEF.
}

\begin{figure} [ht]
    \centering
    \includegraphics[width=1\textwidth]{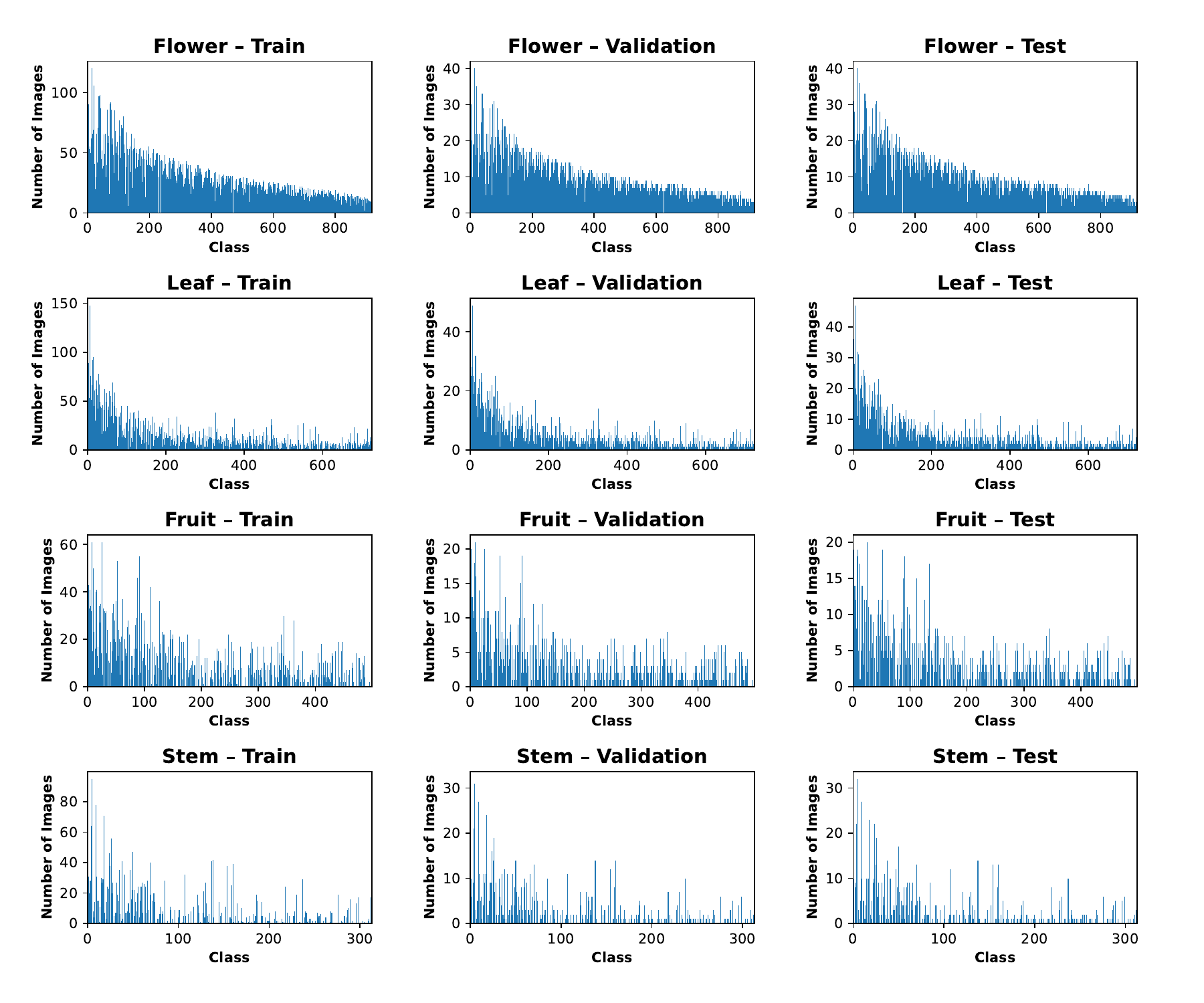}
    \caption{Distribution of Organs across Different Splits}
    \label{fig:splits}
\end{figure}

\begin{enumerate}[label=\textbf{\arabic*.}, ref=\arabic*, wide, labelindent=0pt]

    \item \textbf{Combining the original splits:} The original data splits of PlantCLEF2015 are poorly balanced with respect to plant organs; therefore, we merge them into a unified dataset to generate our own splits subsequently.

    \item \textbf{Collecting metadata:} Initially, we extract image metadata for flowers, leaves, fruits and stems from the dataset, resulting in a total of 67812 records.
    
    \item \textbf{Grouping class observations:} We group the metadata of different organ images by their observation identifiers for each class separately.
    
    \item \textbf{Filtering organ images and classes:} We eliminate specific organ images from a class if their total count in this class is less than 3, as such images cannot be split between training, validation and test sets. Subsequently, we remove any observations that are left empty. Finally, we discard classes with fewer than 3 observations, since we split the data by observations (Section \ref{section:dataset}). This filtering results in 919 classes for flowers, 726 for leaves, 500 for fruits, and 314 for stems, yielding a total of 979 unique classes.

    \item \textbf{Splitting the data:} We split the data by observations; however, achieving balanced splits is challenging as (i) we aim to ensure a specific proporion of observations in each data split---i.e., 60\%:20\%:20\% for training, validation and test sets respectively---to maximize the diversity of data within each split, and,  simultaneously, (ii) we seek to ensure the same proportions of each plant organ's images in the splits to achieve certain split sizes and balance in terms of organs. These objectives are inherently contradictory, as observations contain varying numbers of organ images (Fig. \ref{fig:organ_dist}). Consequently, we frame the splitting process as a constrained optimization problem, which we solve for each class separately to ensure stratification by labels:
    \begin{equation} \label{eq:splitting}
        \begin{split}
            \textrm{minimize\hspace{1em}} & \underbrace{\sum_{s\in S}(||\mathbf{x}_s||_1 - \lambda_sN)^2}_{\text{optimize observations}}+\underbrace{\sum_{o\in O}\sum_{s\in S}(\mathbf{c}_o^T\mathbf{x}_s-\lambda_s||\mathbf{c}_o||_1)^2}_{\text{optimize organ    counts}}\\
            \textrm{subject to\hspace{1em}}
            & \sum_{s\in S} \mathbf{x}_s=\mathbf{1}
        \end{split}
    \end{equation}
    \noindent where $N$ is the number of observations, $S$ is a set of data splits, $O$ is the set of organ types, $\mathbf{c}_o\in \mathbb{N}^N$ is a vector containing the counts of organ $o$ images for each observation, $\mathbf{x}_s\in \{0,1\}^N$ is a decision vector for a split $s$, where the $i$-th element is 1 if the $i$-th observation belongs to split $s$, and $\lambda_s \in [0,1]$ is the desired size of split $s$ ($0.6$ for training and $0.2$ for validation and test).

    After solving this problem, we assign observations to splits according to the resulting vectors $\mathbf{x}_s$. If an organ split in a particular class ends up empty because all images were assigned to other splits, we manually transfer a single image from the largest split to ensure that training, validation, and testing can be performed for that class. This correction is practically expectable when a class has, for example, 3–4 images of the organ. Overall, this procedure produces reasonably balanced splits, as shown in Figure \ref{fig:splits}. Table \ref{tab:splits} summarizes the total counts of observations and organ images across the splits.

    \item \textbf{Saving unimodal datasets:} We save each data split separately for each modality to enable pre-training of unimodal models for MFAS. Each dataset consists of shuffled images and their corresponding labels, with labels mapped to values in the range $[0, 978]$, as the original label values are impractical for use. Each image is converted to RGB, resized to $256\times256$ resolution and encoded in JPEG format. This standardization ensures uniform model input dimensions and reduces computational and I/O load. The choice of a square shape for resizing is justified by the observation that the average aspect ratio of images in the dataset closely approximates 1. During data loading, we also normalize pixel channel values to the range $[-1, 1]$ to meet the input requirements of MobileNetV3Small. \label{step:preprocessing:saving_unimodal_data}

    \item \textbf{Saving multimodal datasets:} In this step, we prepare the data for training the multimodal architecture. To accomplish this, the four modalities must be combined. Each class in the splits encompasses a varying number of images of distinct organs.
        A straightforward approach would entail generating all possible combinations of $\textit{flower}\times\textit{leaf}\times\textit{fruit}\times\textit{stem}$ for each class. However, given the unequal distribution of organ images across classes, this would yield numerous similar tuples, potentially burdening computational resources without significant learning gains.
        Instead, we only generate $N=max(n_{flower},n_{leaf},n_{fruit},n_{stem})$ random image combinations
        per class, where $n_m$ represents the number of images of modality $m$ in the class.
        We create random vectors $\mathbf{x}^{(m)}$ for each modality $m$ 
        by permuting sequences of modality images and then repeating their elements to match the length of $N$. Then, we generate sets of key-value pairs $R_i=\{(m,x^{(m)}_i)|\forall m\}$ for our multimodal dataset, where $i\in\{1,\dots,N\}$. We shuffle these sets and store them together with the corresponding labels to subsequently use them for our multimodal models. If a class entirely lacks a modality $m$, the respective vector $\mathbf{x}^{(m)}$ is not constructed, and $(m, x^{(m)}_i)$ is omitted in records $R_i$ for that class. The formats of stored images and labels follow the specifications in the step \ref{step:preprocessing:saving_unimodal_data}.
\end{enumerate}

\begin{table} [ht]
    \centering
    \begin{tabular}{llll}
        \hline
        \head{Splits:} & \head{Training} & \head{Validation} & \head{Test}\\
        \hline
        \head{Observations} & 16737 (59.21\%) & 5788 (20.48\%) & 5740 (20.31\%)\\
        \head{Flowers} & 21859 (60.09\%) & 7273 (19.99\%) & 7243 (19.91\%)\\
        \head{Leaves} & 9357 (59.84\%) & 3172 (20.29\%) & 3107 (19.87\%)\\
        \head{Fruits} & 5096 (59.97\%) & 1678 (19.75\%) & 1724 (20.29\%)\\
        \head{Stems} & 3196 (59.05\%) & 1119 (20.68\%) & 1097 (20.27\%)\\
        \hline
    \end{tabular}
    \caption{Counts of Observations and Organs in Data Splits}
    \label{tab:splits}
\end{table}

\subsection{Proposed Model Setup and Configuration}

\subsubsection{Unimodal Model Training} \label{section:training_unimodal_models}

Each base MobileNetV3Small model is appended with an intermediate dense layer (with batch normalization) and a classifier layer. Initially, we train only these appended layers using the default Adam optimizer with an exponentially decaying LR and early stopping (ES), after which the best weights are restored based on validation loss. Subsequently, we fine-tune from the top layer down to and including the 4th inverted residual block (IRB) for the leaf and fruit models, and all layers for the flower model. Fine-tuning did not yield performance improvements for the stem model, so it is omitted there. During fine-tuning, ES is delayed until the 40th epoch. Table \ref{tab:unimodal_hp} provides the complete list of unimodal model hyperparameters.

\begin{table} [ht]
    \centering
    \begin{tabular}{lllll}
        \hline
        \head{Models:} & \head{Flower} & \head{Leaf} & \head{Fruit} & \head
        {Stem}\\
        \hline
        \head{Optimizer} & Adam & Adam & Adam & Adam\\
        \head{Initial LR} & $10^{-3}$ & $5^{-4}$ & $5^{-4}$ & $10^{-3}$\\
        \head{LR decay rate} & $0.95$ & $0.95$ & $0.95$ & $0.95$\\
        \head{LR decay steps} & 200 & 200 & 200 & 200\\
        \head{Batch size} & 256 & 256 & 256 & 256\\
        \head{Epochs} & 1000 & 1000 & 1000 & 1000\\
        \head{ES patience} & 10 & 10 & 10 & 10\\
        \head{Classifier L1} & $10^{-5}$ & $0$ & $0$ & $0$\\
        \head{Classifier L2} & $10^{-5}$ & $10^{-3}$ & $10^{-3}$ & $10^{-2}$\\
        \head{Classifier dropout} & 0.4 & 0.1 & 0.2 & 0.0\\
        \head{Intermediate neurons} & 1024 & 1024 & 512 & 1024\\
        \head{Intermediate layer L2} & 0 & $10^{-5}$ & 0 & 0\\
        \hline
        \multicolumn{5}{c}{\head{Fine-tuning:}}\\
        \hline
        \head{Fine-tuning} & True & True & True & False\\
        \head{Initial LR} & $10^{-4}$ & $10^{-4}$ & $8^{-5}$ & ---\\
        \head{ES from epoch} & 40 & 40 & 40 & ---\\
        \head{Fine-tuned layers} & All & Up to 4th IRB & Up to 4th IRB & ---\\
        \hline
        \multicolumn{5}{c}{\head{Retraining on combined training and validation sets:}}\\
        \hline
        \head{ES} & False & False & False & False\\
        \head{Epochs} & 160 & 140 & 120 & 110\\
        \head{Fine-tuning epochs} & 160 & 120 & 180 & ---\\
        \hline
        \multicolumn{5}{l}{Hyperparameters remain unchanged during fine-tuning or retraining on combined}\\
        \multicolumn{5}{l}{datasets unless specified otherwise.}\\
        \hline
    \end{tabular}
    \caption{Hyperparameters of Unimodal Models}
    \label{tab:unimodal_hp}
\end{table}

To minimize the impact of class imbalance, we employ a weighted cross-entropy loss, 
where a weight for each class $c\in C$ is computed according to the formula:
\begin{equation}
    w_c = \frac{N}{|C|} \times N_c
    \label{eq:class_weight}
\end{equation}
\noindent where $N_c$ is the number of instances of a class $c$ within the training data, and $N$ is a total number of instances.

Additionally, we shuffle the data at each epoch and randomly apply image augmentations that have shown to enhance performance, as outlined in Table \ref{tab:unimodal_augmentations}.

\begin{table} [H]
    \centering
    \begin{tabular}{lllll}
        \hline
        \head{Models:} & \head{Flower} & \head{Leaf} & \head{Fruit} & \head
        {Stem}\\
        \hline
        \head{Contrast} & True & False & True & False\\
        \head{Horizontal flip} & True & True & True & True\\
        \head{Vertical flip} & False & True & False & False\\
        \hline
        \multicolumn{5}{l}{Contrast adjustment computes a mean $\mu$ for each image channel $c$ and}\\
        \multicolumn{5}{l}{sets it for each pixel as $(c - \mu)\times k + \mu$, where $k\sim U(0.75,1.25)$.}\\
        \hline
    \end{tabular}
    \caption{Image Augmentations for Unimodal Models}
    \label{tab:unimodal_augmentations}
\end{table}

To fully utilize the available data, we train two versions of each model: one on the training set only, and another on the combined training and validation sets. The first model version is used during MFAS execution (Section \ref{section:mfas_exectution}) and while tuning the final model (Section \ref{section:final_model_training}). Once hyperparameter tuning is complete, the validation set is no longer required, so we merge it with the training set and retrain the models on this combined data for evaluation purposes. Without a validation set, we cannot apply ES based on validation performance; therefore, we determine an appropriate number of training epochs based on the first model version, increasing it slightly to account for the additional data. The number of epochs used for retraining on the combined sets is provided in Table \ref{tab:unimodal_hp}.

\subsubsection{Multimodal Fusion Architecture Search Execution} \label{section:mfas_exectution}
In the MFAS algorithm procedure, each multimodal architecture is trained using the Adam optimizer with a learning rate of 0.001 and a weighted cross-entropy loss, where class weights are calculated according to Equation \ref{eq:class_weight}. The number of neurons per fusion layer is set to 64.
Each architecture is trained for 2 epochs with a batch size of 256. As batching significantly impacts training speed, we cache batches and shuffle them each epoch in buffers of 12, rather than creating new random batches each time. Given the limited number of epochs per architecture during the algorithm's execution, this approach offers a favorable trade-off between maintaining batch randomness and optimizing training speed.
Given the imbalanced nature of the dataset, we score the architectures using the F1$_{macro}$ metric on the validation set. 
Similar to \citet{Perez-Rua_2019}, we initialize our temperature scheduler with $t_{max}=10$, $t_{min}=0.2$, and temperature decay $d=4$.

In this study, we execute the MFAS algorithm 
for 5 iterations, with 4 progression levels each and set the number of sampled architectures to 50. 
It is important to note that exploring all 2592 initial architectures at the start of the algorithm is time-consuming; therefore, they were evaluated in 20 parallel batches. Subsequently, the algorithm was restarted with the obtained results and a pre-trained surrogate, continuing from the second progression level with the 50 best architectures sampled from the first level.

\subsubsection{Final Model Training} \label{section:final_model_training}

To identify the final architecture after executing the MFAS algorithm, we select the 10 most performant configurations and train each for 100 epochs. We employ ES with a patience of 10 and an initial LR of 0.001, which decays exponentially, reaching 95\% of its value every 200 steps. We also apply batch normalization after each fusion layer, while retaining the other hyperparameters used during the architecture search.

Subsequently, we tune the hyperparameters of the best architecture carefully. Initially, the optimal number of neurons per layer is investigated. As suggested by \citet{Perez-Rua_2019}, small weight matrices are utilized during the algorithm to enhance its speed and reduce memory consumption. However, when focusing on a single architecture, this limitation is no longer applicable. 
Next, we test various configurations of LR, its decay, dropout rates, and other regularization techniques to identify the optimal setup. Additionally, we explore fine-tuning this model to potentially enhance its performance further.

After hyperparameter tuning, we combine the training and validation sets, as detailed in Section \ref{section:training_unimodal_models}, and retrain the model. We also use the corresponding versions of the unimodal models trained on the merged sets. To enhance the model's robustness to missing modalities (as discussed in Section \ref{section:dataset}), we implement the multimodal dropout (MD) technique proposed by \citet{Cheerla_2019}. Inspired by a regular dropout, MD involves dropping entire modalities during training. In our case, this means replacing pixel values of an organ image with zeros. This encourages the model to build representations robust to missing modalities. Each modality of a sample is dropped with a similar probability. 
We utilize a low dropout rate of 0.125 because many classes are completely devoid of certain modalities. Given the challenge of validating this technique during training, we train two versions of the final model: one with MD and one without, and subsequently evaluate both.

\subsection{Model Evaluation}
\subsubsection{Establishing Baseline}
The baseline for the proposed model is established as a late fusion of all unimodal models. This serves as a straightforward approach to fusion, enabling the demonstration that the performance difference between the final model and the baseline is attributed to the unique fusion configuration of the final model.

We implement the late fusion
using the averaging strategy \citep{Baltrusaitis_2019}. In this approach, the final prediction for an instance $x$ is calculated according to the formula:
\begin{equation}
    \hat{y}=\mathop{\text{argmax}}_{y \in Y}p_{\text{avg}}(y|x)
\end{equation}
\noindent where $y$ represents a class label from the set of all class labels $Y$, and $p_{\text{avg}}$ denotes the average of probabilities from each model.
Note that in cases where an instance lacks a modality, the corresponding unimodal model is not involved in the prediction generation process.

\subsubsection{Comparison with the Baseline} \label{section:comparison_baseline}

The evaluation of the proposed model relies on standard performance metrics and comparison against the baseline to determine if the
automatic fusion setting of the model leads to improvements. Subsequently, we apply McNemar's test \citep{Dietterich_1998}
to ascertain if there exists a statistically significant difference between the two models.

To be more specific, in this study, we utilize the following performance metrics:
\begin{equation}
    Accuracy=\frac{TP+TN}{TP+TN+FP+FN}
\end{equation}
\begin{equation}
    Precision=\frac{TP}{TP+FP}
\end{equation}
\begin{equation}
    Recall=\frac{TP}{TP+FN}
\end{equation}
\begin{equation}
    F1=2\times\frac{Precision\times Recall}{Precision+Recall}
\end{equation}
\noindent where with respect to each class, $TP$ (true positives) represents the number of correctly identified instances, $TN$ (true negatives) denotes the number of correctly rejected instances, $FP$ (false positives) signifies the number of incorrectly identified instances, and $FN$ (false negatives) indicates the number of incorrectly rejected instances.

For each class, we compute macro averages of each metric, excluding accuracy.
We omit micro averaging since it accumulates all $TP$, $TN$, $FP$, and $FN$ values before calculating the metric, resulting in a value equivalent to accuracy. In contrast, macro averaging calculates each metric for individual classes and then averages these values across all classes. The difference between accuracy and macro-averaged metrics provides insight into the performance variation across classes.
In conjunction with these metrics, top-5 and top-10 accuracies are also calculated. 

\subsubsection{Robustness to Missing Modalities}

To enable a comparison between unimodal models and multimodal models, we also collect the same metrics discussed in Section \ref{section:comparison_baseline} for individual unimodal models.
Moreover, to facilitate an understanding of the robustness capabilities of the proposed model in the absence of certain modalities, we collect the mentioned metrics for the proposed model and the baseline on different subsets of modalities. To evaluate the statistical significance of the differences between these metrics, we apply McNemar's test.

\section{Results} \label{section:results}
This section presents the findings from our comprehensive evaluation of the proposed model. We have structured our results to provide a clear and systematic presentation under various conditions and compared them to an established baseline.
\subsection{Performance of Unimodal Models}

After the training procedure (see Section \ref{section:training_unimodal_models}), the unimodal models have been evaluated. Table \ref{tab:unimodal_metrics} illustrates the resulting performance metrics.
\begin{table} [h]
    \centering
    \begin{tabular} {lllll}
    \hline
    \head{Modalities:} & \head{Flower} & \head{Leaf} & \head{Fruit} & \head{Stem}\\
    \hline
    \head{Accuracy} & 0.6796 & 0.4294 & 0.4936 & 0.2698\\
    \head{Top-5 accuracy} & 0.8597 & 0.6508 & 0.7129 & 0.4622\\
    \head{Top-10 accuracy} & 0.9082 & 0.7293 & 0.7773 & 0.5652\\
    \head{Precision$_{\bm{macro}}$} & 0.5998 & 0.2304 & 0.3215 & 0.1298\\
    \head{Recall$_{\bm{macro}}$} & 0.5794 & 0.2334 & 0.3444 & 0.1449\\
    \head{F1$_{\bm{macro}}$} & 0.5702 & 0.2181 & 0.3143 & 0.1247\\
    \hline
    \end{tabular}
    \caption{Performance Metrics for Unimodal Models}
    \label{tab:unimodal_metrics}
\end{table}

According to Table \ref{tab:unimodal_metrics}: (i) It is evident that the fruit and flower modalities exhibit higher scores across all metrics compared to leaves and especially stems. This is in line with expectations, considering that stems are the least discriminative organs \citep{Nhan_2020}.
(ii) Accuracy demonstrates higher values compared to macro metrics. This discrepancy should be attributed to differences in performance among classes. As mentioned in Section \ref{section:dataset}, the dataset is imbalanced, which has affected the overall performance. 
(iii) Despite that, similar yet lower precision and recall values suggest that models' performance is consistent on more common classes but poorer on less common.
(iv) Relatively high top-$N$ accuracies indicate that the models are typically close to identifying the correct class.

\subsection{Finding Final Architecture}
\begin{figure} [ht]
    \centering
    \resizebox{1\textwidth}{!}{
    \includegraphics{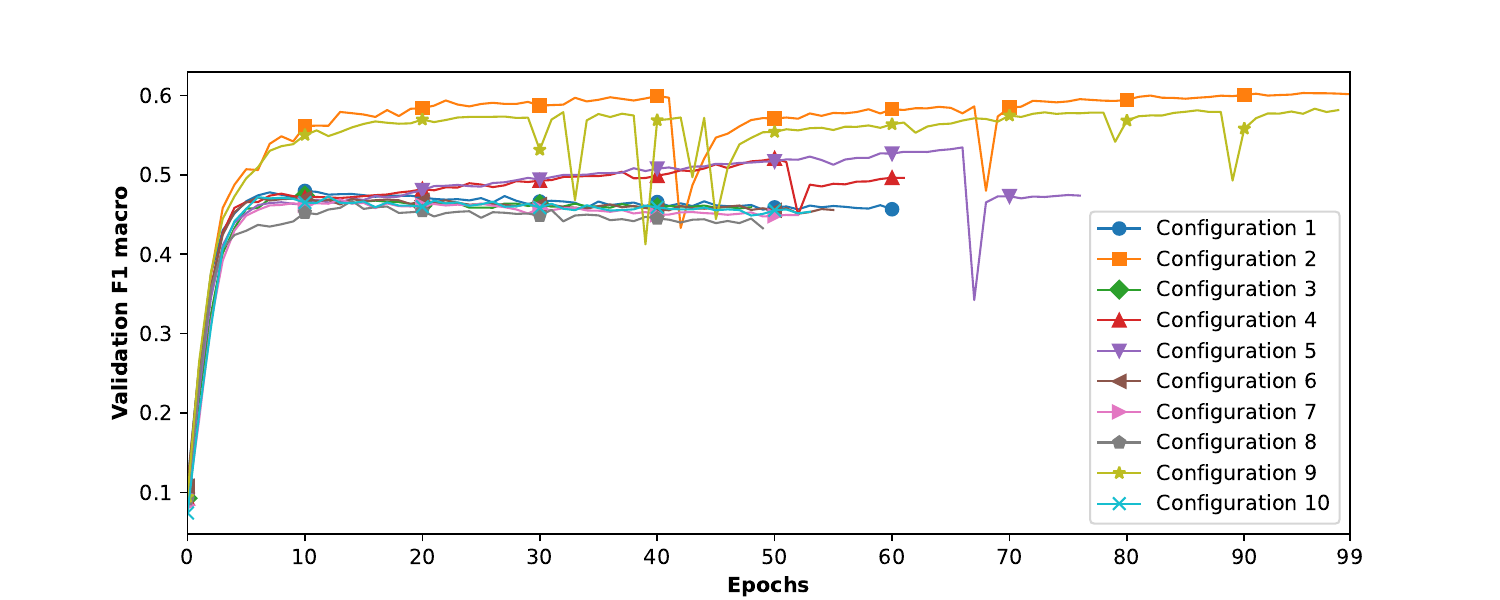}
    }
    \caption{Validation F1$_{macro}$ scores for Sampled Architectures across Epochs}
    \label{fig:best_10_accuracy}
\end{figure}

As a result of the search for an optimal fusion point, we sampled the 10 best configurations, which are presented in Table \ref{tab:best_10_configurations}. Following this, we trained these configurations as described in Section \ref{section:final_model_training}. Figure \ref{fig:best_10_accuracy} illustrates that all architectures achieved approximately 47\%–60\% in validation F1$_{macro}$ scores. Notably, the second-best architecture identified by the MFAS algorithm (ranked 2nd in Table \ref{tab:best_10_configurations}) proved to be the most effective in this test, achieving a validation F1$_{macro}$ of 60.35\%. Consequently, we have selected this architecture for further hyperparameter tuning.

\begin{table} [h]
    \centering
    \resizebox{1\textwidth}{!}{
    \begin{tabular}{llllll}
        \hline
        \head{Nr.} & \head{1st fusion} & \head{2nd fusion} & \head{3rd fusion} & \head{4th fusion} & \head{F1$_{\bm{macro}}$}\\
        \hline
        1 & $ReLU(\flower{B1}, \leaf{A}, \fruit{A}, \stem{O})$ & $\sigma(\flower{B11}, \leaf{B1}, \fruit{B6}, \stem{B6})$ & $\sigma(\flower{B6}, \leaf{B6}, \fruit{B11}, \stem{A})$ & $ReLU(\flower{I}, \leaf{I}, \fruit{I}, \stem{B6})$ & 0.4610 \\
        \textbf{2} & \boldsymbol{$\sigma(\flower{B11}, \leaf{B1}, \fruit{B11}, \stem{O})$} & \boldsymbol{$ReLU(\flower{O}, \leaf{O}, \fruit{B6}, \stem{A})$} & \boldsymbol{$ReLU(\flower{O}, \leaf{B11}, \fruit{B6}, \stem{B11})$} & \boldsymbol{$ReLU(\flower{I}, \leaf{I}, \fruit{O}, \stem{B1})$} & \textbf{0.4561} \\
        3 & $ReLU(\flower{I}, \leaf{I}, \fruit{I}, \stem{B11})$ & --- & --- & --- & 0.4469 \\
        4 & $ReLU(\flower{B11}, \leaf{B6}, \fruit{O}, \stem{B6})$ & $ReLU(\flower{O}, \leaf{B11}, \fruit{A}, \stem{B1})$ & $\sigma(\flower{B6}, \leaf{B6}, \fruit{B11}, \stem{A})$ & $ReLU(\flower{I}, \leaf{I}, \fruit{I}, \stem{B6})$ & 0.4468 \\
        5 & $ReLU(\flower{B11}, \leaf{B6}, \fruit{O}, \stem{B6})$ & $ReLU(\flower{O}, \leaf{B11}, \fruit{A}, \stem{B1})$ & $\sigma(\flower{B6}, \leaf{B11}, \fruit{B11}, \stem{B11})$ & $ReLU(\flower{I}, \leaf{I}, \fruit{I}, \stem{B6})$ & 0.4468 \\
        6 & $ReLU(\flower{B11}, \leaf{B6}, \fruit{O}, \stem{B6})$ & $\sigma(\flower{B11}, \leaf{B1}, \fruit{B6}, \stem{B6})$ & $\sigma(\flower{B6}, \leaf{B6}, \fruit{B11}, \stem{A})$ & $ReLU(\flower{I}, \leaf{I}, \fruit{I}, \stem{B6})$ & 0.4459 \\
        7 & $ReLU(\flower{B1}, \leaf{B1}, \fruit{O}, \stem{B11})$ & $\sigma(\flower{I}, \leaf{B11}, \fruit{B6}, \stem{I})$ & $ReLU(\flower{B1}, \leaf{B1}, \fruit{I}, \stem{B11})$ & $ReLU(\flower{I}, \leaf{I}, \fruit{A}, \stem{B6})$ & 0.4434 \\
        8 & $ReLU(\flower{B11}, \leaf{B6}, \fruit{O}, \stem{B6})$ & $ReLU(\flower{A}, \leaf{A}, \fruit{O}, \stem{A})$ & $\sigma(\flower{B6}, \leaf{B6}, \fruit{B11}, \stem{A})$ & $ReLU(\flower{I}, \leaf{I}, \fruit{I}, \stem{B6})$ & 0.4421 \\
        9 & $\sigma(\flower{B11}, \leaf{B1}, \fruit{B11}, \stem{O})$ & $ReLU(\flower{O}, \leaf{O}, \fruit{B6}, \stem{A})$ & $ReLU(\flower{O}, \leaf{B11}, \fruit{B6}, \stem{B11})$ & $ReLU(\flower{I}, \leaf{O}, \fruit{I}, \stem{B6})$ & 0.4418 \\
        10 & $ReLU(\flower{B6}, \leaf{B6}, \fruit{O}, \stem{B11})$ & $\sigma(\flower{I}, \leaf{B11}, \fruit{B6}, \stem{I})$ & $ReLU(\flower{B1}, \leaf{B1}, \fruit{I}, \stem{B11})$ & $ReLU(\flower{I}, \leaf{I}, \fruit{A}, \stem{B6})$ & 0.4387 \\
        \hline
        \multicolumn{5}{l}{$ReLU$ and $\sigma$ denote activation functions of fusion layers, whereas their parameters denote layers fused from}\\
        \multicolumn{5}{l}{\flower{flower}, \leaf{leaf}, \fruit{fruit} and \stem{stem} unimodal models, respectively. $A$ – first activation, $I$ – intermediate dense layer,}\\
        \multicolumn{5}{l}{$O$ – output layer, $BN$ – $N$-th inverted residual block.}\\
        \multicolumn{5}{l}{The F1$_{macro}$ values were achieved by the models after 2 epochs during the search process (Section \ref{section:mfas_exectution}).}\\
        \hline
    \end{tabular}
    }
    \caption{Best Configurations Sampled by MFAS}
    \label{tab:best_10_configurations}
\end{table}

\subsection{Final Model}

To train the final model, we adjust various hyperparameters of the second architecture from Table \ref{tab:best_10_configurations}. The full list of hyperparameters is presented in Table \ref{tab:final_hp}. Surprisingly, neither full randomization nor augmentations led to improved performance during multimodal model training. As a result, we continue using cached batches shuffled in buffers of 12 each epoch, as done during the MFAS execution (Section \ref{section:mfas_exectution}). Additionally, fine-tuning the model did not enhance generalization, and is therefore omitted.

\begin{table} [ht]
    \centering
    \begin{tabular}{ll}
        \hline
        \head{Hyperparameter} & \head{Value}\\
        \hline
        \head{Optimizer} & Adam\\
        \head{Initial LR} & $5^{-4}$\\
        \head{LR decay rate} & $0.9$\\
        \head{LR decay steps} & 200\\
        \head{Batch size} & 256\\
        \head{Epochs} & 100\\
        \head{ES patience} & 10\\
        \head{Classifier dropout} & 0.4\\
        \head{Fusion dropouts} & [0.0, 0.0, 0.0, 0.4]\\
        \head{Fusion neurons} & [512, 512, 512, 512]\\
        \head{MD} & 0.0 or 0.125\\
        \head{Fine-tuning} & False\\
        \hline
        \multicolumn{2}{c}{\head{Retraining on combined training and validation sets:}}\\
        \hline
        \head{ES} & False\\
        \head{Epochs} & 100\\
        \hline
        \multicolumn{2}{l}{Hyperparameters remain unchanged during retraining on combined}\\
        \multicolumn{2}{l}{datasets unless specified otherwise.}\\
        \hline
    \end{tabular}
    \caption{Hyperparameters of the Final Model}
    \label{tab:final_hp}
\end{table}

\begin{figure} [ht]
    \centering
    \resizebox{1\textwidth}{!}{
    \includegraphics{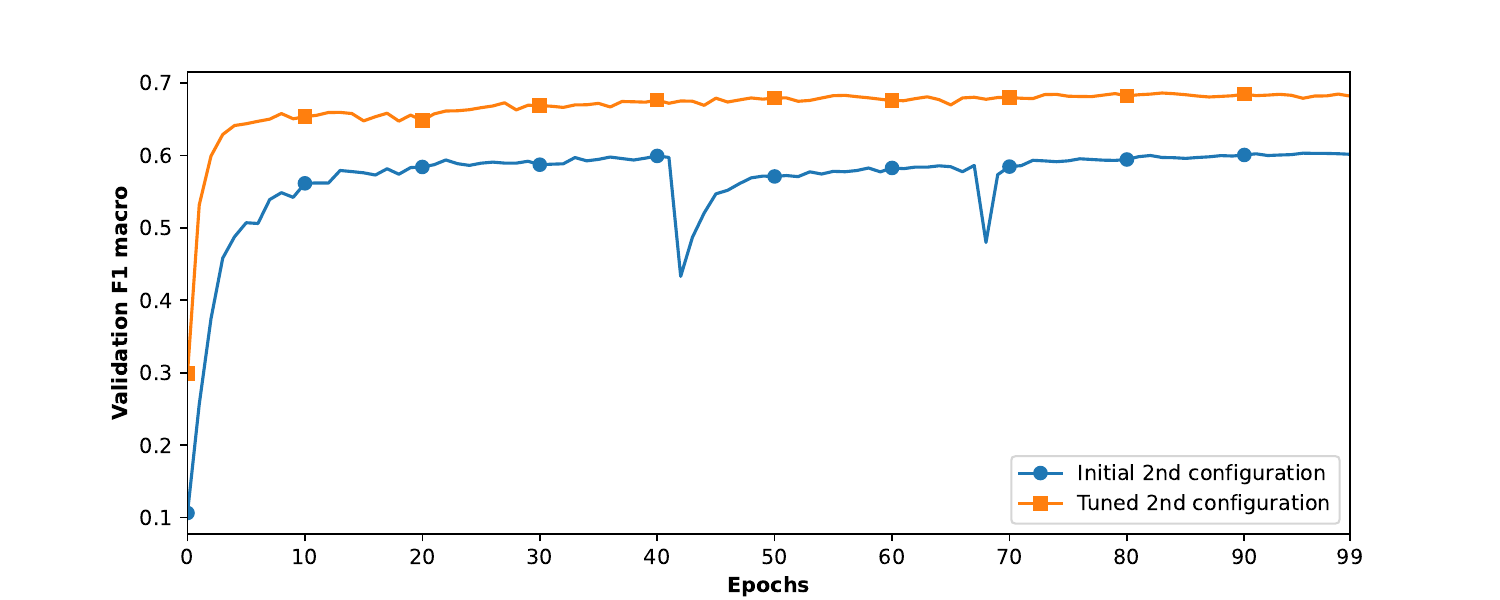}
    }
    \caption{Validation F1$_{macro}$ across Epochs before and after Final Model Hyperparameter Tuning}
    \label{fig:final_validation_f1}
\end{figure}

With all the aforementioned adjustments, the model reached a validation F1$_{macro}$ score of 68.62\%. Figure \ref{fig:final_validation_f1} illustrates the increase in F1$_{macro}$ for this configuration following hyperparameter tuning. To prepare the model for evaluation, we merge the training and validation sets and train two versions of the model---one with MD and one without it, as discussed in Section \ref{section:final_model_training}.

\subsection{Comparison with the Baseline}

We collect the performance metrics for both versions of the proposed model and the baseline on the test set, which are displayed in Table \ref{tab:final_metrics}.

\begin{table} [h]
    \centering
    \begin{tabular} {llll}
    \hline
    \head{} & \head{Proposed Model} & \head{Proposed Model with MD} & \head{Baseline}\\
    \hline
    \head{Accuracy} & 0.8261 & 0.7856 & 0.7228\\
    \head{Top-5 accuracy} & 0.9546 & 0.9219 & 0.8982\\
    \head{Top-10 accuracy} & 0.9754 & 0.9528 & 0.9420\\
    \head{Precision$_{\bm{macro}}$} & 0.7838 & 0.7397 & 0.6652\\
    \head{Recall$_{\bm{macro}}$} & 0.7743 & 0.7130 & 0.6284\\
    \head{F1$_{\bm{macro}}$} & 0.7595 & 0.7024 & 0.6224\\
    \hline
    \end{tabular}
    \caption{Performance Metrics for Proposed Model and Baseline}
    \label{tab:final_metrics}
\end{table}

These results indicate that both versions of the proposed model significantly outperform the baseline across all explored metrics, suggesting that our automated fused model is substantially more effective and reliable for plant classification on the given dataset compared to a straightforward late fusion approach. The improvement is particularly notable in the model without MD, achieving, for instance, +0.1459 in recall and +0.1371 in F1 score. Conversely, a more modest improvement is observed in the model with MD, with, for example, +0.0846 in recall and +0.08 in F1 score. This suggests that, overall, the model without MD generalizes better when evaluated over the entire test set, where modalities in instances are missing only if they are entirely absent in their corresponding classes across all splits. However, in real-world scenarios, the available set of modalities may differ from those represented in certain classes in our dataset. For this reason, we assess the models on subsets of modalities in Section \ref{section:comparison_on_subsets}.

\subsubsection{McNemar's Test Results}

We also conducted McNemar's tests to compare the performance of two model pairs: (i) the proposed model versus the baseline and (ii) the proposed model with MD versus the baseline. The contingency tables of model predictions are provided in Table \ref{tab:mcnemar_results}.

The McNemar's tests yielded statistics of $\chi^2_1 = 534.12$ and $\chi^2_1 = 228.74$, both of which are statistically significant ($p < 0.001$). These results indicate a significant difference in performance metrics between each version of our proposed model and the baseline.

\begin{table} [h]
    \centering
    \begin{tabular} {|l|l|l|l|}
    \hline
    \multicolumn{2}{|c|}{\head{Without MD}} & \multicolumn{2}{c|}{\head{With MD}}\\
    \hline
    $n_{00}$: 1197 & $n_{01}$: 281 & $n_{00}$: 1468 & $n_{01}$: 354\\
    \hline
    $n_{10}$: 1159 & $n_{11}$: 5863 & $n_{10}$: 888 & $n_{11}$: 5790\\
    \hline
    \end{tabular}
    \caption{Contingency Tables of Proposed Model with/without MD and Baseline} 
    \label{tab:mcnemar_results}
\end{table}

\subsection{Comparison on Subsets of Modalities} \label{section:comparison_on_subsets}

One can observe from Table \ref{tab:unimodal_metrics} and Table \ref{tab:final_metrics} that both our proposed model and late fusion significantly outperform each individual unimodal model. 
However, these observations do not fully consider the fact that the unimodal models are evaluated on isolated unimodal datasets, whereas the multimodal model is evaluated on a multimodal dataset containing missing modalities and duplications. To address this and in order to have a fair comparison, we evaluate both our proposed models and the unimodal models on isolated modalities from the multimodal dataset. Furthermore, while cases where a single or all organs are available, represent two extremes, there exists a considerable number of image combinations within the dataset where 2–3 modalities are available. Therefore, it is essential to evaluate the performance of our proposed model and the baseline on all possible subsets of 2–3 modalities. Finally, we assess our models in cases where all modalities are available. Table \ref{tab:final_subsets_metrics} summarizes the results of these evaluations.

\begin{table} [h]
    \centering
    \resizebox{1\textwidth}{!}{
    \begin{tabular} {lllll}
    \hline
    \head{Modalities} & \head{\# of Predictions} & \head{Proposed Model} & \head{Proposed Model with MD} & \head{Baseline}\\
    \hline
    Flower & 8097 & 0.3315** & 0.5512** & 0.5663\\
    Leaf & 6714 & 0.0535** & 0.2133* & 0.2012\\
    Fruit & 4614 & 0.0000** & 0.1259** & 0.2894\\
    Stem & 3531 & 0.0000** & 0.0374** & 0.1203\\
    Flower, Leaf & 6325 & 0.5120** & 0.6285** & 0.5159\\
    Flower, Fruit & 4357 & 0.2987** & 0.5332 & 0.4574\\
    Flower, Stem & 3229 & 0.3384** & 0.5684** & 0.3655\\
    Leaf, Fruit & 4130 & 0.0604** & 0.3665 & 0.3512\\
    Leaf, Stem & 3420 & 0.0674** & 0.3588** & 0.2686\\
    Fruit, Stem & 2675 & 0.0000** & 0.2933** & 0.3370\\
    Flower, Fruit, Leaf & 3887 & 0.5814 & 0.6004** & 0.4787\\
    Flower, Stem, Leaf & 3129 & 0.5641* & 0.6534** & 0.4248\\
    Flower, Fruit, Stem & 2450 & 0.4540** & 0.6114** & 0.4197\\
    Leaf, Fruit, Stem & 2611 & 0.1289** & 0.4921** & 0.4092\\
    Flower, Leaf, Fruit, Stem & 2397 & 0.7197** & 0.6768** & 0.4665\\
    \hline
    \multicolumn{5}{l}{* and ** denote statistical significance, with $p < 0.05$ and $p < 0.001$, respectively, as determined by}\\
    \multicolumn{5}{l}{McNemar's test when compared to the baseline.}\\
    \hline
    \end{tabular}
    }
    \caption{Comparison of F1$_{macro}$ for Proposed and Baseline Models on Subsets of Modalities}
    \label{tab:final_subsets_metrics}
\end{table}

It is important to note that the metrics in Table \ref{tab:final_subsets_metrics} account for the absence of target modalities by excluding instances that miss any of the required modalities. 

The results indicate that the baseline significantly outperforms our proposed model without MD in most cases. The proposed model exhibits superior metrics only in 3 out of 4 scenarios involving 3 modalities and when all modalities are available; however, one of the cases with 3 modalities lacks statistical significance, and another shows lower significance compared to other comparisons. Notably, there are instances where our model achieves a 0\% F1 score, specifically in cases involving only fruits and/or stems, while subsets that include leaves demonstrate higher scores, and those incorporating flowers yield significantly better results than any other subsets. The distribution of scores among the unimodal models suggests a different trend, with the informal notation $F1_{flower} > F1_{fruit} > F1_{leaf} > F1_{stem}$ in unimodal models versus $F1_{flower} > F1_{leaf} > F1_{stem} \geq F1_{fruit}$ in the proposed model, implying that the proposed model may be sensitive to the imbalance in modalities, leaning more towards learning from the more common modalities. Consequently, to enhance performance on subsets with less represented modalities, techniques such as augmenting the data with additional instances of underrepresented modalities or applying weighted loss based on the modalities of a predicted instance should be considered. Overall, the findings suggest that the proposed model without MD is more effective when the majority of modalities are available; however, since it consistently trains with the maximum number of available modalities, this approach ultimately leads to reduced robustness when modalities are missing.

Conversely, the proposed model with MD significantly outperforms the baseline in the majority of cases, indicating the effectiveness of the MD technique in enhancing robustness to missing modalities. Surprisingly, its performance with a single flower modality is very similar to that of the baseline and even better with the leaf modality, despite the baselines, in this cases, being standalone unimodal models specifically designed for individual modalities. This observation highlights the model's ability to resist noise from idle modality-specific parts. Notably, this version of our model surpasses the one without MD in all cases except for the scenario where all modalities are available. This suggests that the robustness to missing modalities afforded by MD comes at a slight cost to performance across all modalities; the model without MD consistently trains on all available modalities simultaneously, facilitating the learning of interconnections between them, while the model with MD rarely encounters the full set of modalities during training.

Overall, our approach outperforms the late fusion of unimodal models. However, if robustness to missing modalities is crucial, it is essential to employ MD or similar techniques. Conversely, to maximize performance when the full set of available modalities is present, the model should consistently train with all the available modalities visible.

\section{Discussion} \label{section:discussion}

\subsection{Accuracy} \label{section:Accuracy}
Our proposed multimodal DL model, which utilizes plant organ images fused through the MFAS algorithm, demonstrates high effectiveness in automating plant classification, achieving an accuracy of 82.61\% and significantly surpassing the defined baseline. Table \ref{tab:study_comparison} indicates that our model achieves the highest accuracy among similar research.

To be more specific, for example, \citet{Ghazi_2017} employed all organs from the PlantCLEF2015 dataset. 
Along with advanced scoring methods, they achieved an accuracy of 80.18\%. 
Similarly, \citet{deLutio_2021} sampled a large dataset comprising 56608 high-quality images of 977 species from the iNaturalist database. 
Utilizing images and spatio-temporal context, their multimodal model achieved accuracies of 79.12\% without satellite imagery and 79.73\% with it. 
Furthermore, \citet{Lee_2018}, also using the PlantCLEF2015 dataset, achieved an accuracy of 68.5\%. \citet{Ge_2016} and \citet{Nguyen_2016} extracted flower species from the PlantCLEF2015 dataset and achieved an accuracy of 52.1\% and 67.45\% respectively. These results are significantly lower than ours, underscoring the effectiveness of our approach.

Extending the related work, we might observe some other multimodal studies achieving better results; however, this is often due to less complex experimental setups. 
For instance, \citet{Zhang_2012} achieved an accuracy of 93.23\%, but on a dataset comprising very small images. Similarly, \citet{Liu_2016} achieved top-1, top-5 and top-10 accuracies of 71.8\%, 91.2\% and 96.4\% without geographical data, and 50\%, 100\% and 100\% with it, respecively. However, their dataset included only 50 species, each with 10 leaf and 10 flower images, collected in controlled conditions. Likewise, \citet{Salve_2018} demonstrated a high GAR score, but their dataset consisted of only 60 species with 10 images each, all collected in laboratory conditions. Furthermore, \citet{Nhan_2020} achieved a maximum 98.8\% accuracy by utilizing all PlantCLEF2015 organs and 91.4–98.0\% depending on model configurations, on only flowers, leaves, fruits, and stems, but they sampled only 50 species from the dataset, each containing all organs, thus not experiencing missing modalities as we did. 
\citet{Seeland_2021} achieved a maximum accuracy of 94.25\% on the PlantCLEF2015 dataset; however, their results were based on a very small selection of images and classes.

In addition to these comparisons, our proposed method demonstrates robustness to missing modalities, as evidenced by its comparison with the baseline in Section \ref{section:comparison_on_subsets}.

\subsection{Computational Cost} \label{section:Computational_Cost}
While our approach involves much \emph{one-time} computational cost during training to identify an optimal architecture, the resulting model is highly efficient during inference. With only 10 million parameters, our model is lightweight and easily deployable to mobile devices. This parameter count could be further reduced by employing fewer fusion layers, if needed.

In comparison, \citet{deLutio_2021} utilize ResNet-50 with 25 million parameters, \citet{Lee_2018} design a custom architecture exceeding 50 million parameters, and \citet{Ghazi_2017} employs VGG and GoogleNet, which contain 138 million and 6.8 million parameters, respectively. Contrary, our model is significantly smaller, more efficient, and suitable for deployment on energy-constrained devices. This underscores the importance of multimodality and an optimal fusion strategy.


\begin{table*}[h]
    \centering
    \resizebox{1\textwidth}{!}{
    \begin{tabular}{llllll}
    \hline
    \head{Study} & \head{Dataset} & \head{Classes} & \head{Modalities} & \head{Method} & \head{Accuracy}\\
    \hline
    \textbf{Proposed method} & \textbf{\cfix{Multimodal-PlantCLEF}} & \textbf{979} & 
    \begin{tabular}{@{}l}
        \textbf{Flower, leaf,} \\
        \textbf{fruit, stem}
    \end{tabular} & 
    \begin{tabular}{@{}l}
        \textbf{MFAS using}\\
        \textbf{MobileNetV3Small}
    \end{tabular} & \textbf{82.61\%} \\
    \hline
    \citet{deLutio_2021} & iNaturalist & 977 &
    \begin{tabular}{@{}l}
        Plant images,\\
        metadata, \\
        satellite imagery\\
    \end{tabular} & ResNet & 79.73\%\\
    \hline
    \citet{Lee_2018} & PlantCLEF2015 & 1000 & Plant organ & 
    \begin{tabular}{@{}l}
        HGO-CNN + \\
        Plant-StructNet
    \end{tabular} & 68.5\% \\
    \hline
    \citet{Ghazi_2017} & PlantCLEF2015 & 1000 & Plant organ & 
    \begin{tabular}{@{}l}
        GoogleNet + \\
        VGG
    \end{tabular} & 80.18 \% \\
    \hline
    \citet{Nguyen_2016} & Modified PlantCLEF2015 & 967 & Flower & GoogleNet & 67.45\% \\
    \hline
    \citet{Ge_2016} & Modified PlantCLEF2015 & 967 & Flower & MixDCNN & 52.1\% \\
    \hline
    \end{tabular}
    }
    \caption{Comparison with Similar Studies}
    \label{tab:study_comparison}
\end{table*}

\subsection{Search Efficiency}
\cfix{
The MFAS algorithm incorporates several optimizations---including surrogate modeling, weight sharing, reduced model complexity, and limiting training to a small number of epochs per configuration---that enable efficient exploration of a large number of architectures. The primary bottleneck in the original procedure occurs during the first iteration, where all generated configurations must be trained to provide data for the surrogate model. To address this, we divided the initial architectures into 20 batches and trained them in parallel. 
}

\cfix{
This approach allowed the algorithm to evaluate up to $2592 + 50 \times 19 = 3542$ unique architectures within a time frame roughly equivalent to training a single multimodal model for $((2592/20)+50\times19)\times2=2159.2$ epochs. Even if the first iteration were performed sequentially, the process would involve $7084$ epochs---a feasible one-time computational cost for identifying an optimal architecture. Moreover, these requirements could be further reduced by decreasing the number of layers considered for fusion, the number of iterations performed or the number of architectures explored.
}

\cfix{
In comparison, manual search for an optimal configuration would require to investigate a search space of size $4.51 \times 10^{13}$, making the MFAS algorithm not only efficient but also practical for large-scale multimodal learning tasks.
}

\cfix{
In this study, we employed MFAS as it is a well-established multimodal NAS algorithm that satisfied our efficiency requirements when compared to alternative methods (e.g., \citep{Xu_2021}). By being the first to apply such methods to plant classification, we establish a benchmark for future research in this area. Future studies in plant classification can investigate other multimodal NAS approaches to further enhance training efficiency.
}

\subsection{Limitations} \label{section:limitations}
\cfix{
In this study, we modify the PlantCLEF2015 dataset, which is a common practice in the literature \citep{Seeland_2021, Nhan_2020, Nguyen_2016, Ge_2016}. This modification was necessary due to the lack of multimodal datasets and the inherent limitations of the original PlantCLEF2015 dataset, which is unsuitable for tasks requiring fixed inputs of specific plant organs. To address this gap and promote comparability and reproducibility in future research, we contribute a preprocessing pipeline that transforms PlantCLEF2015 into Multimodal-PlantCLEF.
}

\section{Conclusions} \label{section:conclusion}

This study has addressed a critical task in agriculture and ecology: plant classification. By proposing a novel approach in this domain---a multimodal DL model utilizing four plant organs automatically fused via the MFAS algorithm, the study has demonstrated high performance, outperforming other state-of-the-art models despite the smaller size of the model. 
This underscores the effectiveness of multimodality and an optimal fusion strategy. 
Moreover, with the MD technique, the model exhibits robustness to missing modalities, even when only a single modality is available. 
\cfix{Additionally, we contributed Multimodal-PlantCLEF, a restructured version of the PlantCLEF2015 dataset tailored for multimodal tasks, to support further research in this area.}
We believe that this proposed approach opens up a promising direction in plant classification research. This highlights the need for further exploration through the development of more sophisticated algorithms capable of handling larger numbers of species, thereby unlocking the full potential of this approach. 
A future research could also consider incorporating a multimodal fusion of vision transformers instead of CNNs for plant classification.

\section*{Acknowledgements}
The computations were enabled by resources provided by the National Academic Infrastructure for Supercomputing in Sweden (NAISS) at Chalmers Centre for Computational Science and Engineering (C3SE) partially funded by the Swedish Research Council through grant agreement no. 2022-06725.

\balance

\bibliographystyle{elsarticle-harv} 
\bibliography{ref.bib}

\end{document}